# Feature Based Fuzzy Rule Base Design for Image Extraction

Koushik Mondal
Indian Institute of Science Education and Research, Pune, India
gemkousk@ieee.org

Paramartha Dutta
DCSS, Visva Bharati University, Santiniketan, West Bengal, India
paramartha@ieee.org

Siddhartha Bhattacharyya
Department of CS & IT
RCC IIT, Kolkata, WB, India
siddhartha.bhattacharyya@gmail.com

*Abstract*— In the recent advancement of multimedia technologies, it becomes a major concern of detecting visual attention regions in the field of image processing. The popularity of the terminal devices in a heterogeneous environment of the multimedia technology gives us enough scope for the betterment of image visualization. Although there exist numerous methods, feature based image extraction becomes a popular one in the field of image processing. The objective of image segmentation is the domain-independent partition of the image into a set of regions, which are visually distinct and uniform with respect to some property, such as grey level, texture or colour. Segmentation and subsequent extraction can be considered the first step and key issue in object recognition, scene understanding and image analysis. Its application area encompasses mobile devices, industrial quality control, medical appliances, robot navigation, geophysical exploration, military applications, etc. In all these areas, the quality of the final results depends largely on the quality of the preprocessing work. Most of the times, acquiring spurious-free preprocessing data requires a lot of application cum mathematical intensive background works. We propose a feature based fuzzy rule guided novel technique that is functionally devoid of any external intervention during execution. Experimental results suggest that this approach is an efficient one in comparison to different other techniques extensively addressed in literature. In order to justify the supremacy of performance of our proposed technique in respect of its competitors, we take recourse to effective metrics like Mean Squared Error (MSE), Mean Absolute Error (MAE) and Peak Signal to Noise Ratio (PSNR).

I. INTRODUCTION

In traditional computing methodology, the prime considerations are precision, certainty, and rigor. Its aim is to exploit the tolerance due to imprecision, uncertainty, approximate reasoning, and partial truth in order to achieve tractability, robustness, and low-cost solutions. The guiding principle is to devise methods of computation that lead to an acceptable solution at low cost by seeking for an approximate solution to an imprecisely/precisely formulated problem. This, in essence, is the guiding principle of soft computing [1]. Soft computing is a consortium of methodologies that works synergistically and provides in one form or another flexible information processing capability for handling real life ambiguous situations. The theory of fuzzy logic [2] offers an extensive mathematical framework to capture the uncertainties associated with human cognitive processes, such as thinking and reasoning. The conventional approaches of knowledge representation lack the means for representing the meaning of fuzzy concepts. As a consequence, the approaches based on first order logic and classical probability theory do not provide an adequate conceptual framework for dealing with the representation of commonsense knowledge, since such knowledge is by its nature both lexically imprecise and non-categorical. Fuzzy Logic is usually regarded as a formal way to describe how human beings perceive everyday real life concepts. In fuzzy image processing, fuzzy set theory [3] is applied quite convincingly. Fuzzy image processing depends upon membership values [4], rule-base and inference engine. Unlike classical logic systems, Fuzzy Logic (FL) system aims at modeling the imprecise modes of reasoning, which is the human ability to make a rational decision when information is uncertain and imprecise. FL starts with the concept of a fuzzy set. A fuzzy set is a set without a crisp, clearly defined boundary. It can contain elements with only a partial degree of membership. Membership criteria are not precisely defined for most classes of objects normally encountered in the real world. A fuzzy set is characterized by a membership function, that takes values in the interval [0, 1], such that the nearer the value to unity, the higher is the membership grade. The uncertainty in image extraction and subsequent segmentation from noise affected scene effectively handled by Fuzzy Logic. According to [5], fuzzy approaches for image segmentation can be categorized into four classes: (i) segmentation via thresholding, (ii) segmentation via clustering, (iii) supervised segmentation and (iv) rule based segmentation. Among these categories, rule based segmentation approach is able to take advantage of application dependent heuristic knowledge and model them in the form of fuzzy rule base. In our case, the heuristic knowledge gathers by the process of existing threshold based segmentation methods that helped us to build the rule base. Thresholding is a simple shape extraction technique. If it is assumed that the shape to be extracted is defined by its brightness, then thresholding an image at that brightness level should find the shape. Thresholding is clearly sensitive to change in intensity. If the image intensity changes then so does the perceived brightness/color of the target shape. Unless the threshold level can be arranged to adapt to the change in brightness level, any thresholding technique is expected to fail. Its attraction is simplicity since thresholding does not require much computational effort. If the intensity level changes in a linear fashion, then histogram equalization results in an image that does not vary in respect of intensity. Unfortunately, the result of histogram equalization is sensitive to noise. It can affect the resulting image quite dramatically and this will help us to determine the minute changes in the original images clearly. Image segmentation and subsequent extraction from noise-affected scene happen to be crucial phase of image processing. The extraction task transforms rich content of images into various content features. Feature extraction is the process of generating features to be used in the selection and classification tasks. Feature selection reduces the number of features provided to the classification task. Those features which are likely to assist in discrimination are selected and used in the classification task. Features not selected are discarded [6]. In this paper, we consider different threshold values, red, green, blue, mean and standard deviation as features. We use these features as input for fuzzy rule base system for generating appropriate membership functions. After the features are extracted, a suitable classifier need be chosen. A number of classifiers are used and each classifier is found suitable to classify a particular kind of feature vectors depending upon their characteristics. Fuzzy Rule Base Systems (FRBS) techniques are one of the powerful tools in classification domain. They are widely used because they make us able to handle imprecise, noisy or incomplete information often present in many classification problems. FRBS have the ability to build a linguistic model interpretable to the users and to mix different information, such as the information coming from the feature extractions, mathematical models and empirical measures [7][8]. The most important advantage of FRBS is that they have a high interpretability of the output model. The structures present inside the features represent the information in an organized manner so that the relationship among the variables in the classification process can be identified. The issue of choosing the features to be extracted should be guided by the following concerns:

- The features should carry enough information about the image and should not require any domain-specific knowledge for their extraction.
- They should be easy to compute in order that the approach is feasible for a large image collection.
- They should relate well with the human perceptual characteristics since users will finally determine the suitability of the retrieved images.

This chapter is presented in the following manner. Sections II and III introduce fuzzy image representation and pixel and feature based image segmentation respectively. Our proposed algorithmic approach is described in section IV followed by result and analysis in section V. Section VI provides conclusions and future direction of the paper.

## II. FUZZY IMAGE REPRESENTATION

Fuzzy image processing is the collection of all approaches that understand, represent and process the images, their segments and features as fuzzy sets. First the fuzzy representation of a spatial domain image, a gray tone image I of dimension M x N, with L gray levels, is considered as a two dimensional array of fuzzy singleton sets,

$$I = \{(x_{mn}, \mu_{mn}(x)) | (m,n) \in [(0,0),(M-1,N-1)]\} \quad (1)$$

where, each pixel is characterized by the intensity value $x_{mn}$ and its grade of possessing some membership $\mu_{mn}$ ($0 \leq \mu_{mn} \leq 1$). Each membership value corresponds to the spatial intensity value of image in the range [0, L-1]. Fuzzy property can be expressed in terms of a membership function. For the transformation of a grayscale image I in the range of [0, L-1] into the fuzzy representation in the interval [0, 1], a Gaussian membership function of the form

$$\mu(x_{mn}) = e^{[-(x_{max}-x_{mn})^2/2f_h^2]} \quad (2)$$

has been suggested in [9] and contains a single fuzzifier, $f_h$. $x_{max}$ and $x_{mn}$ are the maximum and (m,n)th gray values respectively. This same function can be used for histogram based membership representation. Following two equations are used to calculate the value of PSNR

$$MSE = \frac{\sum_{M,N}[I_1(m,n) - I_2(m,n)]^2}{M*N}$$

$$PSNR = 10*\log_{10}(\frac{R^2}{MSE}) \quad (3)$$

$R$ is the maximum fluctuation in the input image data type. For example, if the input image has a double-precision floating-point data type, then $R$ is 1. If it has an 8-bit unsigned integer data type, $R$ is 255, etc. Different approaches exist for computing the PSNR of a color image. Because the human eye is most sensitive to intensity information, compute the PSNR for color images by converting the image to a color space that separates the intensity channel with weighted average of R, G, and B. G is given the most weight as because the human eye perceives it most easily. A fuzzy system comprises five basic elements, as shown in Figure 1. A fuzzifier is responsible for mapping the crisp inputs from the system into the corresponding fuzzy set representation. The second element is the knowledge base, which incorporates the required knowledge about the system in the form of fuzzy If-Then rules.

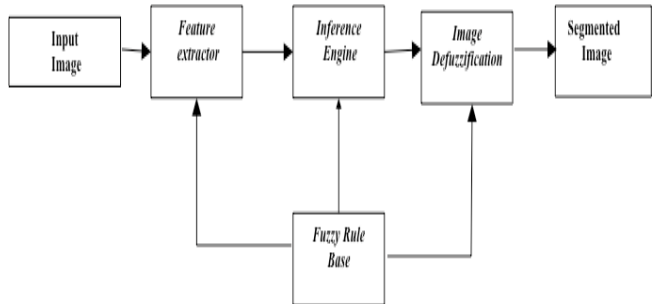

Figure 1: Schematic diagram of Fuzzy Image Processing

The rules are governed by the relationships between the inputs and the way they combine to produce the desired output. The third element is the fuzzy model, which is the group of fuzzy sets describing each of the input and output variables. The fuzzy sets partition the universe of discourse of a given input or output variable into a group of overlapping fuzzy sets. The fourth element is the fuzzy inference system, which is the reasoning process through which the fuzzified inputs are used to activate the relevant rules. The last component is the defuzzifier, which is the mechanism by which the fuzzy input set is converted back into a single output value or control parameter.

The fuzzification and defuzzification steps are particularly important because of absence of any fuzzy hardware at our disposition. Therefore, the coding of image data (fuzzification) and decoding of the results (defuzzification) are steps indispensible that make possible to process images with fuzzy techniques.

### III. PIXEL AND FEATURE BASED SEGMENTATION

An image is collection pixels which have feature vectors. The features of a pixel could be gray level for gray scale images, or red, green, blue levels for color images. The artificial features: mean, noise etc., could also be added into feature vector as additional but relevant components. In image processing field, the similarity measure of two pixels has been generally assessed so far by means of Euclidian distance in color space. On the other hand, Wuerger et al. [10] showed that perceptual color proximity in color space is not Euclidean in nature. That means that distance information in Euclidean color space is not adequate for similarity judgment. Recently, Color Category Map based on Fuzzy Feature Contrast Model was constructed by Seaborn et al [11]. To achieve fine-grain segmentation at the pixel level, we must be able to define features on a per-pixel basis. Different features are often taken into account in conjunction with color information to achieve better segmentation results than possible with just color alone. To enable fine-grained segmentation, we must be able to segment down to the level of individual pixels for which feature values must be extracted on a per-pixel basis. For color information, this is straightforward since the three color components defining the pixel color intensity are usually used as its color features. For groups of pixels, the color features are represented by the mean and standard deviation of the components color values. In this article, we propose a novel extraction method with the help of the above mentioned features viz. color components, threshold values, mean and standard deviation. We used 17 many well known threshold techniques for initial segmentation and those methods also helped us to build membership functions at the later stage. The values that we received from different threshold techniques will form some region of interests in the input image. This will help us to find some similarity regions. The concept of similarity is very important as it provides the stepping stone for organizing the world into categories. In daily life, we often come across situations where we have to distinguish similar groups or we have to classify some similar objects. Similarity measure thus becomes an important tool to decide the similarity degree between two groups or between two objects. Psychologists have developed two main categorization models namely, Similarity-based categorization and Rule-based categorization [12]. The similarity must involve rules and there must be rules for determining a similarity value for any pair of concepts as Hampton [13] and others working on the rule based categorization. Rule-based color similarity was also applied by Demirci et al. [14] for color image segmentation.

### IV. OUR APPROACH

It is interesting that all the thresholding methods invariably performed poorly for at least one or two instances. It was also observed that any single algorithm could not be successful for all noisy image types, even in a single application domain. To obtain the robustness of the thresholding method, we explored the combination of more than one thresholding algorithm based on the conjecture that they could be complementary to each other.

The algorithm for our proposed work is as follows:

Step 1. Read a noisy image as input

Step 2. Identify the Region of Interests (ROI)s of the image by different thresholding values

Step 3. Extract the image information in terms of pixel attributes and threshold values for future use.

Step 4. Construct the different membership envelops of the input image.

Step 5. Generate fuzzy rules based on the numerical data obtained from the input image corrupted by noise. The fuzzy

rule generation consists following steps:
   a. Discern Input and Output spaces into fuzzy regions
   b. Generate fuzzy rules from the given data
   c. Map the threshold values obtained from different methods in the corresponding fuzzy region
   d. Create a combined fuzzy rule base. Determine a mapping on the basis of this combined fuzzy rule base.

Step 6. Approximate the value obtained in Step 5.

Step 7. Display the image constructed thus.

The combination of thresholding algorithms can be done at the feature level or at the decision level. At the feature level, we use, for example, some averaging operation on the maximum values obtained from individual algorithms whereas on the decision level, we apply fusion of the foreground-background decisions, for example, by taking the majority decision. Thus it will help us on creating membership envelops in the proposed system. The main power of fuzzy image processing lies in the effective use of the middle step (modification of membership values). Fuzzy membership functions are defined for each term set of each linguistic variable in the rules.

## V. RESULTS AND ANALYSIS

The goal of this paper is to describe a generic system using a Mamdani rule base. Specifically, we are modeling the relationship among the images, its extracted counterpart and the fuzzy rule base system using as many as 17 well known thresholding methods. In pattern recognition and image processing, feature extraction is a special form of dimensionality reduction. Feature extraction is a general term for methods for constructing combinations of the variables, but still describes the data sufficiently accurately. All images of different categories can be distinguished via their homogeneousness or feature characteristics. All the thresholding methods are generally based on the characteristics of one or some features, which will help us to build an adaptive mechanism guided by some already established methods. In [15][16][17], authors successfully presented fuzzy rule based feature selection approach for gray scale images. Now, in the present context it is extended for baboon color image (256X256) with different Gaussian noise levels. The comparison between all the existing techniques and our proposed technique to extract images are listed in Table 1. The comparisons among different PSNR values are depicted in Figure 2. As PSNR value is depends upon MSE, we only produce the comparison table for PSNR. In this paper, we take proper care about how well a Mamdani rule base can be put to effective use for modeling the system, using rules that have high correctness. For Gaussian noise, the corrupted image, subsequent result obtained by well-known methods and proposed feature based fuzzy rule base method is depicted in Figures 2, 3 and 4 respectively.

## VI. CONCLUSION

The main features and advantages of this approach are:

1. It provides us a general method to combine measured numerical information into a common framework - a combined fuzzy rule base that theoretically entertains both numerical and linguistic information

2. It is a simple and straightforward single pass buildup procedure and hence is devoid of any time consuming iterative training as it happens in a comparable neural network or in a neuro-fuzzy approach

3. There is a lot of freedom in choosing the membership domains in the said design. In fact, this happens to be one of the fundamental challenges

4. This can be viewed as very general integrated fuzzy system devoid of any model for a wide range of image processing tasks where "devoid of model" means no mathematical model is required for formulation of the problem; "integrated" means the system integrates all the reported threshold values that are integrated with the systems for finding ROIs and that can help to design adaptive fuzzy regions. Moreover, "Fuzzy" denotes the fuzziness induced into the system due to linguistic fuzzy rules, fuzziness of data, etc.

The objective criterion for assessing the quality of images relies on the result of computing some of the following statistical error based methods dependent on pixel intensity difference. Overall image mean absolute error (MAE), overall image mean square error (MSE), signal-to-noise ratio (SNR) and peak signal-to-noise ratio (PSNR). The smaller the MAE (or MSE) or the larger the SNR (or PSNR) is, the better is the quality of the signal. It is fast and repeatable.

There is no universal theory on image segmentation and extraction as yet that may be universally applicable in all types of images. This is because image segmentation is subjective in nature and suffers from uncertainty. All the existing image extraction approaches are, in the main, ad hoc. They are strongly application specific. In other words, there are no general algorithms vis-à-vis color spaces that are uniformly good for all color images. An image segmentation and extraction problem is fundamentally one of psychophysical perception and essential to supplement any mathematical solutions by a priori knowledge about the image. The fuzzy set theory has attracted more and more attention in the area of image processing because of its inherent capability of handling uncertainty. Fuzzy set theory provides us with a suitable tool, which can represent the uncertainties arising in image segmentation and can model the relevant cognitive activity of the human beings. Fuzzy operators, properties, mathematics, inference rules have found more and more applications in image segmentation.

The more important advantage of a fuzzy methodology lies in that the fuzzy membership function provides a natural means to model the uncertainty prevalent in an image scene. Subsequently, fuzzy segmentation results can be utilized in feature extraction and object recognition phases of image processing and subsequent computer vision.

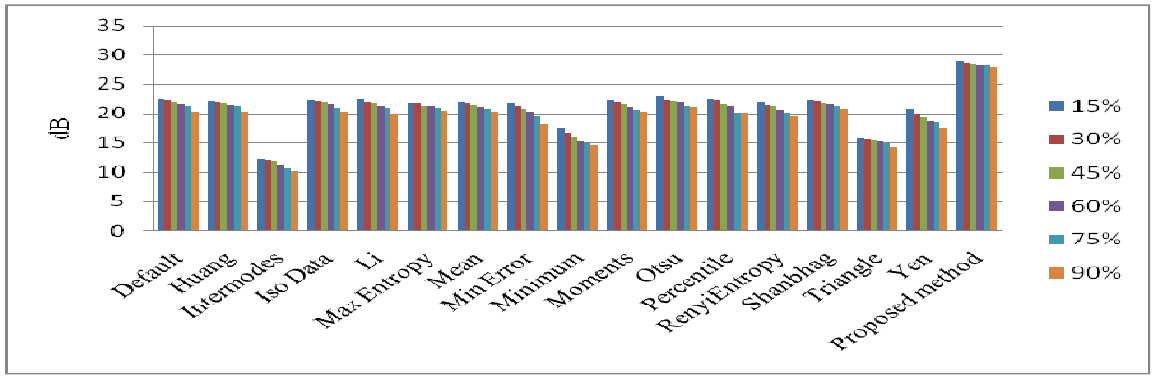

Figure 2: PSNR comparisons of existing techniques and proposed method

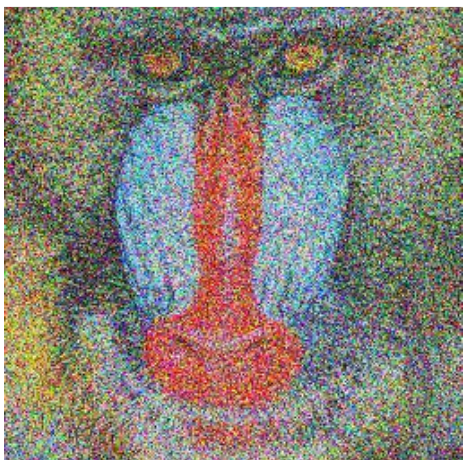

Figure 3: Noisy Image

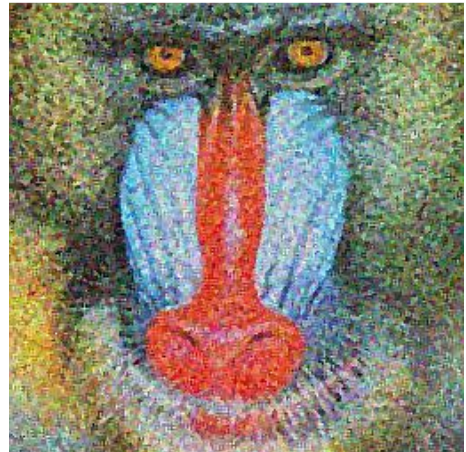

Figure 4: Extracted image by proposed method

TABLE I. COMPARISON OF DIFFERENT THRESHOLDING METHOD AND OUR PROPOSED METHOD

| PSNR Calculation for Gaussian Noise | | | | | | |
|---|---|---|---|---|---|---|
| Threshold methods/Sigma | 15 | 30 | 45 | 60 | 75 | 90 |
| Default | 22.3495 | 22.3035 | 21.9651 | 21.4635 | 21.2171 | 20.2691 |
| Huang | 22.1012 | 21.9167 | 21.838 | 21.3761 | 21.1913 | 20.2827 |
| Intermodes | 12.1539 | 11.9702 | 11.8364 | 11.2701 | 10.6743 | 10.2126 |
| IsoData | 22.2514 | 22.1526 | 21.8981 | 21.4983 | 20.9488 | 20.2725 |
| Li | 22.4745 | 22.0039 | 21.6372 | 21.1526 | 20.8981 | 20.055 |
| Max Entropy | 21.8505 | 21.6211 | 21.2608 | 21.1607 | 20.8364 | 20.4452 |
| Mean | 22.0061 | 21.8187 | 21.376 | 21.1012 | 20.7888 | 20.2452 |
| Min Error | 21.6761 | 21.1951 | 20.6977 | 20.3536 | 19.4598 | 18.2452 |
| Minimum | 17.6406 | 16.6143 | 15.9809 | 15.136 | 14.9167 | 14.6211 |
| Moments | 22.2199 | 21.9488 | 21.4598 | 21.0039 | 20.6211 | 20.2742 |
| Otsu | 22.9488 | 22.2447 | 22.0787 | 21.9167 | 21.2341 | 21.0039 |
| Percentile | 22.4916 | 22.2213 | 21.5802 | 21.2745 | 20.2267 | 20.1325 |
| RenyiEntropy | 21.9618 | 21.3838 | 21.1979 | 20.6492 | 20.2126 | 19.4021 |
| Shanbhag | 22.315 | 22.0904 | 21.7888 | 21.4497 | 21.1931 | 20.7211 |
| Triangle | 15.7136 | 15.6011 | 15.3188 | 15.1825 | 14.85 | 14.2163 |
| Yen | 20.7909 | 19.9472 | 19.2452 | 18.6861 | 18.4514 | 17.4313 |
| Proposed method | 28.9962 | 28.6861 | 28.4983 | 28.3035 | 28.1607 | 27.8981 |

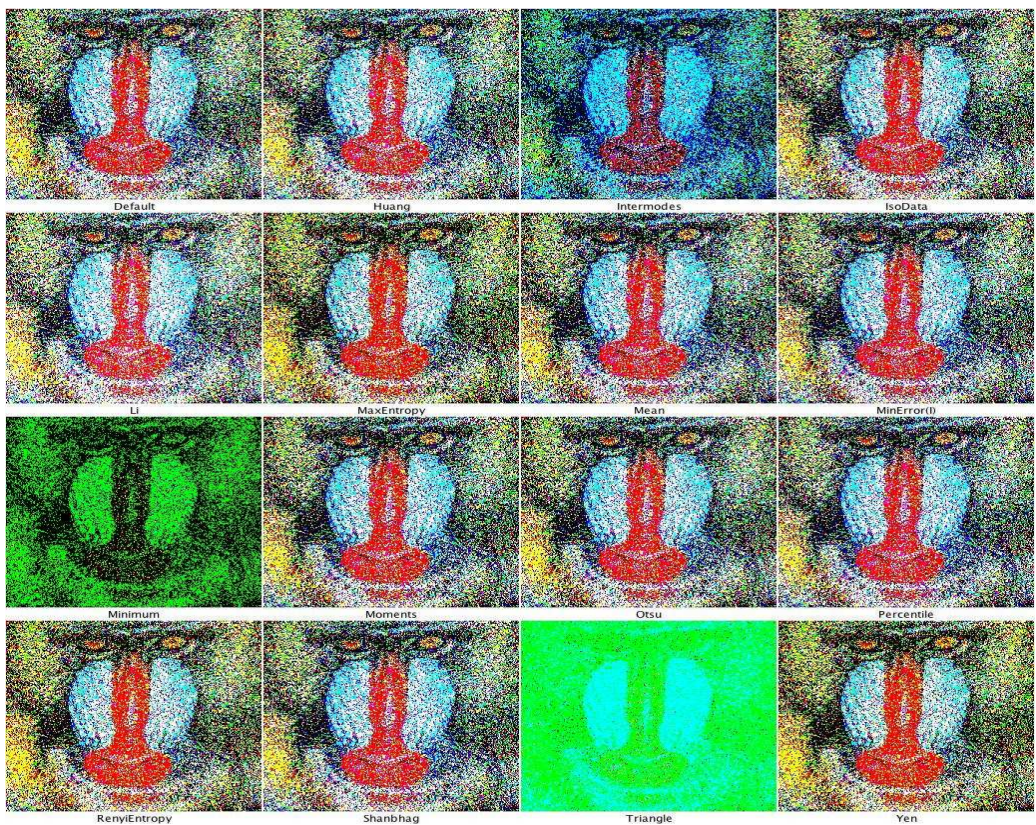

Figure 5: PSNR comparisons of existing techniques and proposed method